\documentclass[11pt]{article}

\usepackage[preprint]{acl}

\usepackage{times}
\usepackage{latexsym}
\usepackage{booktabs}
\usepackage{amsmath}
\usepackage{amsfonts}
\usepackage[T1]{fontenc}

\usepackage[utf8]{inputenc}
\usepackage{microtype}

\usepackage{inconsolata}

\usepackage{graphicx}

\usepackage{cleveref}
\usepackage{xspace}
\usepackage[dvipsnames]{xcolor}
\usepackage{soul}
\usepackage{dsfont}

\newcommand{\defn}[1]{\textbf{#1}}

\title{Beyond Static and Linear:\\ What Attention Constraints Best Fit Human Reading Times?}

\author{
  \textbf{Lanni Bu\textsuperscript{1}},
  \textbf{Xiulin Yang\textsuperscript{2}},
  \textbf{Christian Clark\textsuperscript{3}},
  \textbf{Alex Warstadt\textsuperscript{4}},
  \textbf{Ethan Gotlieb Wilcox\textsuperscript{2}},
\\
\\
  \textsuperscript
  {1}The Ohio State University,
  \textsuperscript
  {2}Georgetown University,
  \textsuperscript{3}New York University,
  \textsuperscript{4}UC San Diego,
\\
  \small{
    \textbf{Correspondence:} \href{mailto:lb1437@georgetown.edu }{bu.138@osu.edu }
  }
}

\begin{document}
\maketitle

\begin{abstract}
Transformer-based language models are widely used as models of human language processing, yet their attention mechanisms allow lossless access to the full preceding context, unlike the limited memory systems of humans.
We hypothesize that installing memory constraints into Transformers' attention mechanisms can improve their fit to human behavioral data.
While previous work has explored individual constraints in isolation, we conduct a systematic comparison of multiple attention-based memory mechanisms across different model sizes and training corpora, evaluating both psychometric predictive power for human reading times and grammatical competence.
We additionally compare static constraints, in which the constraint strength is fixed throughout training, to dynamic memory curricula.
We find that constraints that are sensitive to the content of intervening tokens achieve the highest alignment with human reading times.
However, no constraint leads to a statistically significant improvement over vanilla baselines.
In addition, we observe a dissociation between psychometric fit and grammatical competence under dynamic memory curricula, suggesting that Transformers cannot serve as a one-size-fits-all cognitive model. \footnote{We will release our code upon publication.}

\end{abstract}

\section{Introduction}

One longstanding hypothesis in cognitive science is that constraints on cognition lead to more robust and generalizable cognitive processes \citep{hertwig2003more, newport1990maturational,elman1993learning}. Recently, Artificial Neural Network (ANN)--based language modeling has been used as a testbed for this proposal, specifically in the domain of linguistic cognition. In this line of work, limitations akin to memory limitations in humans are installed into ANN language models, and the resulting models are evaluated on their ability to predict human linguistic behaviors. Recent contributions have found that limitations in Transformers' attention mechanism \citep{clark2025linear, mita2025developmentally, kuribayashi2022context,de-varda-marelli-2024-locally}, as well as representational capacity \citep{timkey2023language, xu-etal-2026-memory}, have led to gains in predicting human grammatical and sentence processing behaviors.
Furthermore, one recent study has found that a curriculum of constraint relaxation over training improves cognitive fit above static model architecture \citep{mita2025developmentally}, in line with well-studied theories from the cognitive science literature \citep{newport1990maturational}.

While these results are exciting, they provide only a limited empirical picture. In particular, previous work has tested only a couple of memory constraints: $N$-gram-like cutoffs \citep{kuribayashi2022context} or distance-based decay \citep{clark2025linear,de-varda-marelli-2024-locally}. However, there are many other theories about how memory constrains language processing (e.g., intervener-based interference; \citealp{lewis2005activation}). These alternative theories remain untested in this paradigm. Furthermore, existing studies have explored a limited range of models and datasets, making it unclear whether the observed improvements in cognitive fit are generalizable or specific to particular training and modeling choices. The same limitations apply to work on dynamic memory curricula: only one type of constraint has been tested under a dynamic training regime, and it is unknown how other memory limitations interact with dynamic versus static training schedules.

In this paper, we test multiple attention-based memory constraints drawn from the pre-existing machine learning literature that have been argued to improve generic language modeling \citep{press2021train, lin2025forgetting, tan2025scaling}.
Even though these models were not originally proposed as cognitive models, we select them because each, we argue, has a plausible cognitive interpretation (see \Cref{sec:cognitive_interp}).\footnote{For a similar, concurrent study investigating distance-based mechanisms, see \citet{madhyastha2026working}.}
 
We vary between two model sizes and three pretraining corpora to assess the generalizability of our findings across different modeling conditions, and train models for a fixed number of epochs.
In addition, we compare static memory constraints, in which the constraint strength remains fixed throughout training, to dynamic memory constraints, in which the constraint changes over the course of training. 
We evaluate the model's ability to predict psycholinguistic behavioral data across six reading time corpora, following well-established methods \citep{smith2013effect, goodkind2018predictive, wilcox2020predictive}. Our interpretation is that if adding a constraint into the model improves its fit to human behavior, then this provides simulational evidence in favor of the corresponding psycholinguistic theory.

We find that constraints that limit models' attention based on the content of intervening material achieve the highest alignment with human reading times.
However, no constraint leads to statistically significant improvement over vanilla baseline models, suggesting that previous results may not extend to new experimental settings.
For dynamic memory curricula, we observe a dissociation between psychometric predictive power and grammatical competence: static models better predict human reading times, but some dynamic models show stronger performance on grammatical benchmarks.

\section{Background}

\subsection{Theories of Memory Limitations in Language Processing}

Cognitive constraints are limitations on the speed, accuracy, and type of mental operation that minds can perform. 
Several influential works have argued that, rather than a mere hindrance, cognitive constraints are advantageous, insofar as they lead to better generalization and robustness \citep{hertwig2003more, newport1990maturational, elman1993learning}.
One domain where this has been argued to be operative is in linguistic cognition (e.g., \citealp{Hitczenko2014CognitiveLI}, \citealp{yang2016price}), with many works investigating the role of memory limitations in language representation \citep{hawkins2004efficiency,gibson2019efficiency} and language processing \citep{gibson1998linguistic,just1992capacity,futrell2020lossy,lewis2005activation,lewis2006computational}.
In this area, one current frontier involves proposing and testing algorithmic-level theories, in the sense of \citet{marr2010vision}, for how memory is limited, by using LMs as platforms for instantiating concrete operationalizations of a given theory.
We review several such theories below, grouping them into two distinct accounts for \emph{how} memory functions during online processing---\emph{distance}-based and \emph{content}-based.
We broadly frame each account within a cue-based retrieval framework of language processing \citep{lewis2005activation,lewis2006computational}.

\paragraph{Distance-based Accounts}

One factor affecting retrieval success in the cue-based retrieval framework is activation decay: memory representations lose activation as a function of time since encoding \citep{lewis2005activation,lewis2006computational}. This can be operationalized as the distance between a target and the retrieval cue. 
A linear decay function has been adopted in computational models of sentence processing, including lossy-context surprisal \citep{futrell2020lossy}.
A related proposal is that memory representations become unavailable beyond a fixed distance, rather than decaying gradually \citep{cowan2001magical}.
Recent computational support for the distance-based account comes from work showing that a linear recency bias in Transformer attention improves a model's ability to predict human reading times \citep{clark2025linear}, that locally biased Transformers with exponential decay show similar improvements \citep{de-varda-marelli-2024-locally}, and that restricting language models' context access to an $N$-gram-like window results in a similar predictive improvement \citep{kuribayashi2022context}.

\paragraph{Content-based Accounts} 
Under content-based accounts, retrieval difficulty arises not from temporal or linear distance, but from featural overlap between the retrieval target and other items in memory, typically those that (linearly) intervene between the two. \citet{van2003distinguishing} proposed a cue-based parsing framework and provided experimental evidence that processing difficulty stems from cue overload at the point of retrieval, rather than from simple temporal decay. \citet{lewis2005activation} subsequently formalized this framework in a computational model built within the ACT-R cognitive architecture, in which sentence comprehension consists of a series of cue-driven retrieval operations from working memory. When retrieval cues fail to uniquely identify the target the result is similarity-based interference, which can lead to processing disruptions and slowdowns. 
The predictions of this account have been validated in controlled psycholinguistic experiments across a range of syntactic phenomena including featural agreement \citep{wagers2009agreement} and thematic binding \citep{van2006retrieval,van2007interference} among others. (See \citealp{jager2017similarity} for a meta-analysis of this literature.)
Despite this strong experimental foundation, content-based interference has not yet been systematically tested within the Transformer-based cognitive modeling paradigm.


\subsection{Cognitive Modeling with Transformer Attention} \label{sec:cognitive_interp}

We look for evidence for the above theories by implementing the constraining mechanism they propose inside a computational learning model and comparing it to an unconstrained baseline model.
We evaluate the cognitive fit of our models based on their predictive power for modeling human syntactic judgements and reading time data. 
For a fair comparison, we make all modifications to the same baseline model, in our case, a vanilla Transformer \citep{vaswani2017attention}.
We focus on decoder-only Transformers because of their previous success as cognitive models, and the parallels between their attention mechanism and cue-based memory retrieval \citep{ryu2021accounting}.

Below, we provide formal definitions of the attention mechanisms we use, along with a cognitive interpretation for each. 
One important caveat is that none of these models was originally designed specifically as a cognitive model of human language processing. Rather, each has been selected because it has been found to improve performance on language modeling tasks, and because we argue it can be mapped onto a plausible cognitive theory of memory limitation.


\newcommand{\mcolor}{Blue}
\newcommand{\dll}{{\textcolor{\mcolor}{\ensuremath{\Delta_{llh}}}}\xspace}
\newcommand{\alibi}{{\textcolor{black}{\text{ALiBi}}}\xspace}

\newcommand{\queries}{{\textcolor{\mcolor}{\ensuremath{\mathbf{q}}}}\xspace}
\newcommand{\keys}{{\textcolor{\mcolor}{\ensuremath{\mathbf{k}}}}\xspace}
\renewcommand{\i}{{\textcolor{\mcolor}{\ensuremath{i}}}\xspace}
\renewcommand{\j}{{\textcolor{\mcolor}{\ensuremath{j}}}\xspace}
\renewcommand{\k}{{\textcolor{\mcolor}{\ensuremath{k}}}\xspace}
\renewcommand{\dim}{{\textcolor{\mcolor}{\ensuremath{d}}}\xspace}

\newcommand{\weights}{{\textcolor{\mcolor}{\ensuremath{\mathbf{w}}}}\xspace}
\newcommand{\bias}{{\textcolor{\mcolor}{\ensuremath{b}}}\xspace}
\newcommand{\embeds}{{\textcolor{\mcolor}{\ensuremath{\mathbf{x}}}}\xspace}

\newcommand{\slopet}{{\textcolor{\mcolor}{\ensuremath{m_t}}}\xspace}
\newcommand{\slopezero}{{\textcolor{\mcolor}{\ensuremath{m_0}}}\xspace}
\newcommand{\lamt}{{\textcolor{\mcolor}{\ensuremath{\lambda_t}}}\xspace}
\newcommand{\lamzero}{{\textcolor{\mcolor}{\ensuremath{\lambda_0}}}\xspace}
\newcommand{\decayrate}{{\textcolor{\mcolor}{\ensuremath{r}}}\xspace}
\newcommand{\windowt}{{\textcolor{\mcolor}{\ensuremath{N_t}}}\xspace}
\newcommand{\contextlen}{{\textcolor{\mcolor}{\ensuremath{C}}}\xspace}

\newcommand{\headslope}{{\textcolor{\mcolor}{\ensuremath{m_h}}}\xspace}

\paragraph{Vanilla Attention \citep{vaswani2017attention}}

The attention $a_{\i,\j}$ between a cue token at index \i, and a target token at index \j, determines to what extent the former is copied over to the latter, changing its in-context representation.
In its vanilla implementation, the attention between two tokens is
determined solely by the token-wise key and query vectors,
$\keys, \queries \in \mathbb{R}^{\dim}$:
\begin{align}
a_{\i\j} =
\mathrm{softmax}_{\j}
\left(
\frac{\queries{\i}^\top \keys_{\j}}
{\sqrt{\dim}}
\right)
\end{align}

\paragraph{Attention with Linear Biases (ALiBi; \citealp{press2021train})} This modification adds a linear penalty to attention scores based on the distance between tokens:
\begin{align}
    a_{\i\j} =
    \mathrm{softmax}_{\j}
    \left(
    \frac{\queries_{\i}^\top
    \keys_{\j}}
    {\sqrt{\dim}} -
    \headslope(\i-\j)
    \right)
\end{align}

\noindent Where \headslope is a hard-coded head-specific scalar.
As ALiBi adds a linear bias to the attention score, we use it to instantiate a distance-based memory constraint.

\paragraph{$N$-gram attention} Also called \emph{sliding window attention}, this restricts the model's context window to a fixed number of preceding tokens at each layer, implementing a cutoff:
%

{\small
\begin{align}
    a_{\i\j} = \mathrm{softmax}_{\j}\left(
    \frac{\queries_{\i}^\top \keys_{\j}}{\sqrt{\dim}} + M_{\i\j}
    \right)
\end{align}
where
\[
M_{\i\j} =
\begin{cases}
    0, & \i - \j < N,\\
    -\infty, & \i - \j \geq N.
\end{cases}
\]
}

Note that $N$ provides a cutoff only \emph{per layer}. 
Attention can flow between tokens separated by more than $N$ positions by propagating information through multiple layers.
Therefore, this attention should be thought of as a soft, rather than a hard, cutoff.
However, the relatively shallow models we use in our experiments ($2$ and $4$ layers) ensure relatively local information sharing.
We take $N$-gram attention as instantiating distance-based memory constraints.

\paragraph{Forgetting Gate (FoX; \citealp{lin2025forgetting})} 
This modification also adds a bias term to each attention score.
However, the bias term is sensitive to the identity of interveners, \k:
\begin{align}
    a_{\i\j} = \mathrm{softmax}_{\j} \left(\frac{\queries_{\i}^\top \keys_{\j}}{\sqrt{\dim}} + \sum_{\k=\j+1}^{\i} \log f_{\k}\right)
\end{align}

\noindent where $f_{\k} = \sigma(\weights_f^{\top}\embeds_{\k} + \bias_f)$ for a given embedding, \embeds, layer-specific learnable weights, \weights and bias, \bias.
Attention in this model is no longer solely a function of the identities of words at index \i and \j but also the identities of words that occur in between them.
Because, in cue-based retrieval theories, linearly intervening words with certain features are hypothesized to interfere with search in memory, we take FoX as an instantiation of content-based memory constraints.

\paragraph{Stick-Breaking attention \citep{tan2025scaling}}

This modification down-weights attention scores based on the presence of high-attention intervening tokens:
%
%
\begin{align}
    a_{\i\j} = g_{\i\j} \prod_{\j<\k<\i}(1 - g_{\i\k}); {\tiny g_{\i\j} = \sigma\left(\frac{\queries{\i}^\top \keys_{\j}}{\sqrt{\dim}}\right)}
\end{align}

\noindent Like FoX, attention decrease is not tied to token positions, although it may be influenced by distance indirectly, insofar as tokens that are more distal from each other will have more interveners.
Unlike FoX, the model does not learn additional parameters that modulate the extent of attention decrease; rather, it gates information using existing \keys and \queries parameters.
We take stick-breaking attention as another case of content-based memory constraint.

\subsection{Dynamic vs. Static Memory Constraints}
\label{sec:dynamic}

The attention mechanisms described above can be applied as fixed constraints throughout training. 
However, an alternative approach is to vary the strength of these constraints over the course of training, motivated by developmental theories of language acquisition. 
The \defn{Less is More hypothesis} \citep{newport1990maturational} proposes that the reason cognitive limitations are useful is that they are embedded within dynamic maturational trajectories. When young, infants' attention and memory constraints force them to attend to certain types of local relationships in their input data. This supports early linguistic analysis, including learning word boundaries, word-to-word transitions, and chunking. This early analysis enables later generalization, which onsets as cognitive capacity matures \citep{elman1993learning}.\looseness=-1

Support for this hypothesis comes from both behavioral studies, which have shown that adults learn artificial languages more effectively when first exposed to simplified input \citep{kersten2001less}, and that children, but not adults, tend to regularize inconsistent input, suggesting that their limited memory capacity leads to more systematic linguistic representations \citep{hawkins2004efficiency,kam2009getting}, as well as computational simulations demonstrating that restricting a model's processing capacity improves its ability to acquire morphological rules \citep{goldowsky1993modeling}.
Recently, in Transformer LMs, \citet{mita2025developmentally} found that slowly relaxing a distance-based attention constraint led to improved grammatical competence.
However, their setup tested only one model configuration and did not assess cognitive fit to human real-time language processing data. 

We test the Less is More hypothesis by embedding our attention-based memory constraints within a dynamic memory curriculum. For each model, we start with a fully constrained attention mechanism and gradually relax it over the course of training (\emph{Less-to-More}), following the setup proposed by \citet{mita2025developmentally}. We contrast each case with an inverse memory curriculum (\emph{More-to-Less}), in which we slowly impose the constraint over the course of training. If the Less is More hypothesis holds for our neural learners, we expect the Less-to-More curriculum to improve cognitive fit over the More-to-Less curriculum. We operationalize the dynamic curricula for the two attention mechanisms that involve adding additional bias terms.
Both of our dynamic implementations add a weight to this bias term, which is slowly reduced or strengthened over the course of training.
More information about our specific implementation of each architecture can be found in \Cref{app:dynamic_models}.\footnote{We explored a curriculum-based version of the $N$-gram attention model, but found that step-changes in $N$ lead to instability during training. We leave a dynamic version of this model and the stick-breaking attention for future work.}

\section{Methods}

\subsection{Language Model Training Data}
\label{sec:training_data}
We train our models on three corpora of varying sizes. The first two are drawn from BabyLM \citep{warstadt2023findings}, and are designed to approximate the quantity and quality of linguistic input available to children during language acquisition: BabyLM-10M (10 million tokens) and BabyLM-100M (100 million tokens). We additionally train models on a 2-billion token subset of the Pile \citep{gao2020pile}, a large-scale web-based English corpus. The inclusion of multiple training corpora allows us to assess whether the effects of memory constraints on cognitive fit are robust across different data scales and compositions.

\subsection{Models and Training}
We use a decoder-only Transformer architecture based on OPT \citep{zhang2022opt} as the basis for all our models. For each attention variant, we train models in two sizes: a 2-layer and a 4-layer configuration. All models are trained from scratch on each of the three corpora described in \Cref{sec:training_data}. 
We train versions for all the attention mechanisms described in \Cref{sec:cognitive_interp}, and include $N \in \{2, 3, 5\}$ for our $N$-gram models. Each model is trained with three random seeds (42, 43, 44), and all reported results are averaged across seeds.
For dynamic models, we implement the curricula described in \Cref{app:dynamic_models} for ALiBi and the Forgetting Transformer. Each dynamic model is trained under both the Less-to-More and More-to-Less conditions.
For static models, we use the final checkpoint after 10 epochs of training. 
For dynamic models, we save and evaluate all checkpoints to examine the trajectory of cognitive fit across training.

\subsection{Human Sentence Processing Datasets}
\label{sec:data}
We evaluate our models on six English reading time corpora using two experimental paradigms. For self-paced reading (SPR), we use the Brown \citep{smith2013effect}, Natural Stories \citep{futrell2018natural}, and UCL \citep{frank2013reading} datasets. For eye-tracking (ET), we use the Dundee \citep{kennedy2013frequency}, GECO \citep{cop2017presenting}, and Provo \citep{luke2018provo} datasets.
Brown comprises SPR data from 35 participants reading 13 passages (7,188 words); Natural Stories, SPR data from 181 participants reading 10 naturalistic stories (10,256 words); UCL, SPR data from 117 participants and ET fixation durations from 48 participants reading isolated sentences (4,957 words); GECO, fixation durations from 14 monolingual participants reading a full-length novel (56,441 words); Dundee, fixation durations from 10 participants reading 67 newspaper editorials (51,501 words); and Provo, fixation durations from 84 participants reading 55 short passages (2,746 words).

For SPR corpora, we exclude reading times for sentence-initial and sentence-final words, as well as observations shorter than 50 ms or longer than 1000 ms. For ET corpora, we analyze first-pass and go-past durations, excluding data points for unfixated words, words following saccades longer than four words, and words at sentence and document boundaries. Each corpus is split into fit (50\%), exploratory (25\%), and held-out (25\%) partitions. Regression models are trained on the fit partition,
Results are reported for the exploratory portion, except for statistical tests, which are conducted on the held-out portion.

\begin{figure*}[t]
    \centering
    \includegraphics[width=\linewidth]{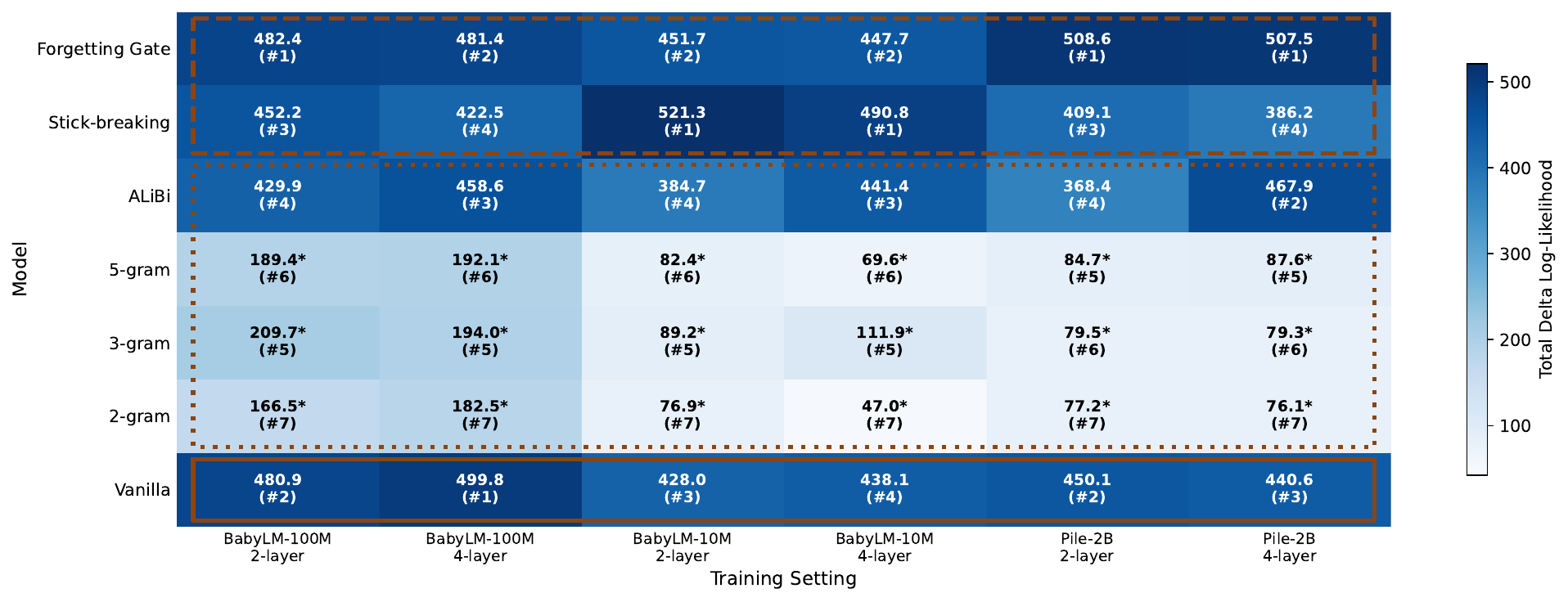}
    \vspace{-0.8cm}
    \caption{Total \dll and rank across training settings. Outline boxes indicate mechanism type: solid = baseline, dashed = interference-based, dotted = distance-based. * indicates a significant difference from vanilla in the second-level mixed-effects model ($p < .05$) within each setting. Text color differences are for visual clarity only.}
    \label{fig:static_heatmap}
\end{figure*}

\subsection{Measuring Human Cognitive Fit}
\label{sec:cognitive_fit}

\newcommand{\word}{{\textcolor{\mcolor}{\ensuremath{w}}}\xspace}
\newcommand{\words}{{\textcolor{\mcolor}{\ensuremath{\mathbf{w}}}}\xspace}
\newcommand{\surp}{{\textcolor{\mcolor}{\ensuremath{\iota}}}\xspace}
\newcommand{\testdata}{{\textcolor{\mcolor}{\ensuremath{\mathcal{T}}}}\xspace}
\newcommand{\llh}{{\textcolor{\mcolor}{\ensuremath{llh}}}\xspace}

We measure human cognitive fit using Delta Log Likelihood or \dll \citep{goodkind2018predictive, wilcox2023testing}, which quantifies the improvement in predicting human reading times when model-derived surprisal is added as a predictor to a regression model with baseline predictors. The surprisal of a word $\word_t$ in context $\words_{<t}$ is defined as:
\begin{align}
    \surp_t(\word_t) = -\log_2 p(\word_t \mid \words_{<t})
\end{align}
where $p(\word_t \mid \words_{<t})$ is estimated using our language models.

For each corpus, we fit two linear mixed-effects models \citep{bates2015fitting} to reading times: one model with just baseline predictors, and one model with baseline predictors plus surprisal. The baseline model includes word length in characters, sentence position, unigram surprisal estimated using KenLM \citep{heafield2013scalable} on the OpenWebText Corpus \citep{gokaslan2019openwebtext}, and, to account for spillover effects, the unigram surprisal, length, and reading time of the previous word. For ET corpora, the baseline additionally includes whether the previous word was fixated. All models include by-subject random intercepts and by-subject random slopes for every predictor, by-word random intercepts, and random intercepts for sentences nested within subjects. The target model extends the baseline by adding model-derived surprisal at the current and previous word, both as fixed effects and as by-subject random slopes. Models are fit on the fit partition using maximum likelihood estimation, and \dll is evaluated on the exploratory partition, \testdata. If $\llh()$ gives a model's log-likelihood for a single test point, then: 

\begin{align}
    \dll = \sum_{\word \in \testdata} \llh_{\text{target}}(\word) - \llh_{\text{baseline}}(\word)
\end{align}

Models whose surprisal leads to higher values of \dll can be said to have higher cognitive fit, or psychological accuracy \citep{frank2011insensitivity, fossum2012sequential}.

\subsection{Measuring Grammatical Competence}
\label{sec:blimp}

We additionally evaluate our models on the BLiMP benchmark \citep{warstadt2020blimp}, which tests grammatical knowledge across 12 linguistic phenomena including subject-verb agreement, argument structure, binding, and filler-gap dependencies. BLiMP consists of minimal pairs of sentences that differ in grammaticality, and a model is scored based on whether it assigns higher probability to the grammatical sentence. We report overall accuracy averaged across all 12 categories.
Our BLiMP scores are not high compared to at-scale LLMs, but well within the typical range reported for models of our size and data scale (for example, against comparable models from the BabyLM challenge; \citealp{hu-etal-2024-findings}, Fig. 2).


\begin{figure}[t]
    \centering
    \includegraphics[width=0.8\columnwidth]{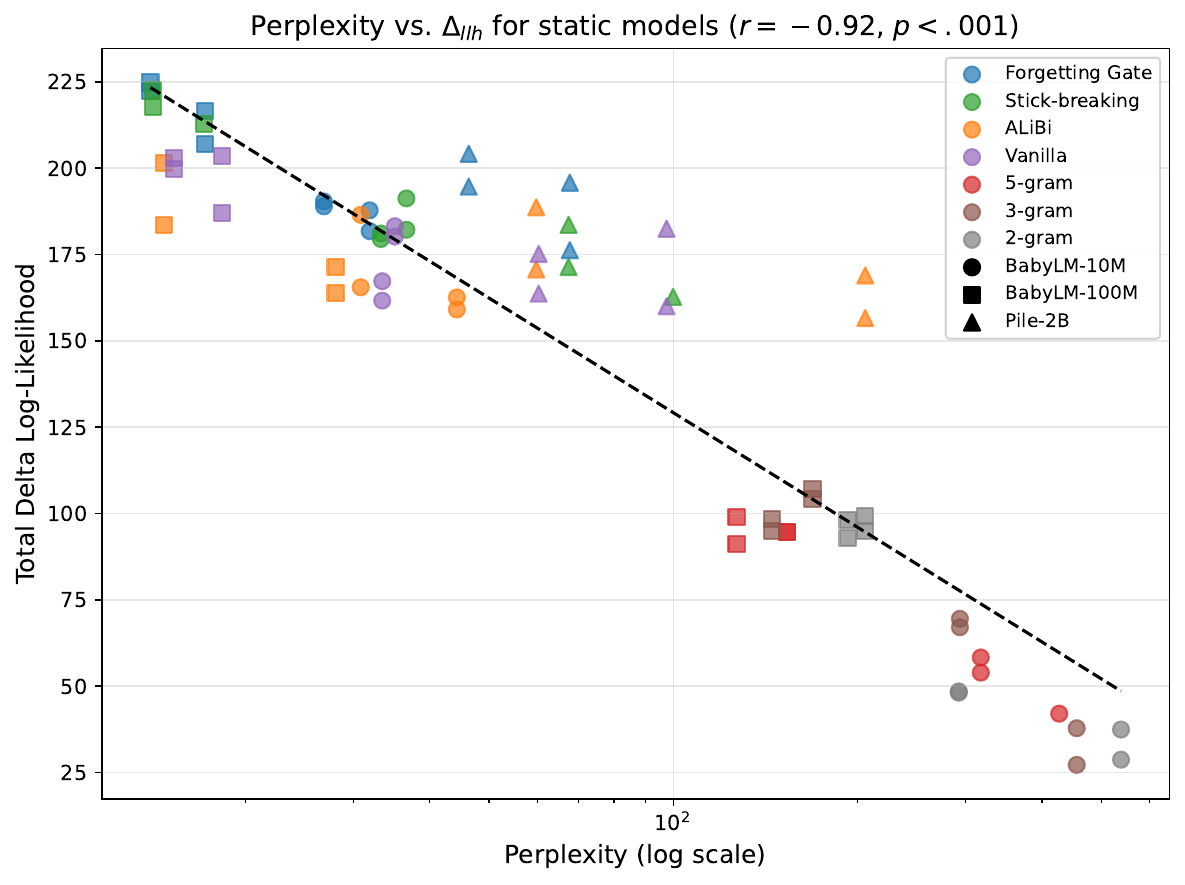}  
    \caption{Perplexity vs.\ \dll for static models. Black dashed line is the line of best fit.}
    \label{fig:static_ppl_dll}      
\end{figure}

\begin{figure*}[t]
    \centering
    \includegraphics[width=0.8\textwidth]{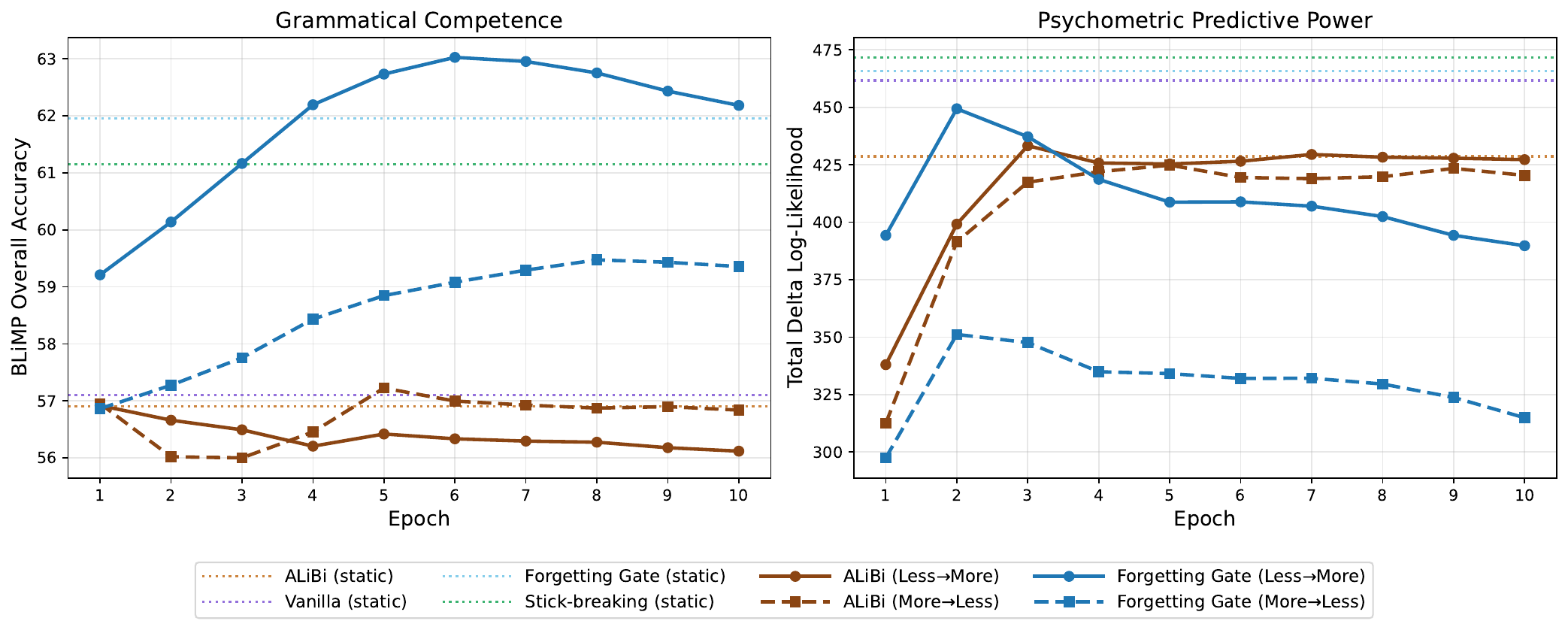}
    \caption{Average \dll (right) and BLiMP scores (left) across training epochs, averaged over four BabyLM training configurations. Line type indicates training regime: solid = Less-to-More, dashed = More-to-Less, dotted = static baseline (final epoch only). Color indicates model.}
    \label{fig:dynamic}
\end{figure*}

\section{Results}

\subsection{Exp. 1: Static Memory Constraints}

\Cref{fig:static_heatmap} reports the summed \dll for each static model across all reading time datasets and training settings.
We test whether \dll significantly differs from the vanilla baseline using a linear mixed-effects model, fit separately for each training setting, with per-word \dll as the dependent variable, attention variant as a fixed effect (vanilla as the reference level), and by-participant and by-document random intercepts and slopes for attention variant, pooling across seeds.
When the model was singular, we refit it with uncorrelated random effects.
$p$-values were computed using Satterthwaite's approximation \citep{kuznetsova2017lmertest}, with significance assessed at $p<0.05$.
Our top-line result is that no model achieves statistically significant improvement over the vanilla baseline in any setting.
This suggests that the gain in cognitive modeling from adding in attention constraints is minimal compared to the overall variance in the data.
While this prevents us from drawing strong conclusions, several trends did emerge in the pattern of results, which are worth noting:
First, we find that content-based attention mechanisms achieve the highest alignment with human reading times; a content-based mechanism ranks first in five out of six training configurations.
In contrast, distance-based mechanisms show weaker results. Although previous work \citep{clark2025linear} found that ALiBi produces higher \dll than the vanilla model, we find that this improvement is sensitive to model size: ALiBi exceeds the vanilla model only with 4-layer models (BabyLM-10M and Pile-2B) and falls below it in all 2-layer settings.
The $N$-gram models achieve \dll values that are significantly lower than the vanilla baseline in all settings.
While these results provide some suggestive evidence in favor of content-based memory constraints, the main conclusion we draw is that experimental settings, including model layer and training data, can have a large impact on the results of reading-time analyses, with different experimental settings leading to different rankings among our models.

\Cref{fig:static_ppl_dll} shows the relationship between perplexity and \dll across all static models. Lower perplexity is generally associated with higher \dll ($r = -.92$, $p < .001$), which is in line with previous work investigating models at this data scale \citep{wilcox2020predictive}. 

\paragraph{Follow-up Analysis: Continued Training}

One potential confound is that our models are trained for only 10 epochs, which is slightly under the typical range for small cognitive models.
(For example, baselines in the BabyLM challenge are typically trained for 20 epochs.)
To assess the impact of training budget, we conducted a follow-up analysis in which we doubled training to 20 epochs.
We find that BLiMP accuracy and \dll remain similar in both settings, suggesting that underfitting is not impacting the main picture of our results.
This experiment is presented in detail in \Cref{app:ep20}.

\subsection{Exp. 2: Dynamic Memory Constraints}

\Cref{fig:dynamic} shows the average \dll and BLiMP scores across training epochs for dynamic models.
Static models, for which only final-epoch scores were recorded, are shown as horizontal bars.
Results indicate averages across all four BabyLM training configurations; for by-dataset and by-model size breakdowns see \Cref{app:dynamic_results_breakdown}.

We observe a partial dissociation between grammatical competence and psychometric predictive power. On the BLiMP benchmark (left panel), the Less-to-More FoX Transformer improves steadily over the first half of training and outperforms all static models from epoch 4 onward.
The More-to-Less FoX Transformer also improves over epochs but remains below its static counterpart, and dynamic ALiBi shows no BLiMP gains over static ALiBi.
However, when looking at \dll (right panel), static models consistently outperform dynamic models. The static FoX Transformer and Stick-breaking attention achieve the highest \dll values, and no dynamic model exceeds its static counterpart by the final epoch.
Within dynamic regimes, a curriculum in which constraints are gradually relaxed (Less-to-More) yields better psychometric fit and grammatical competence, but only for the FoX model.

Overall, this dissociation suggests that the goals of modeling real-time human sentence processing and grammatical acquisition may require different approaches. Static memory constraints, which aim to produce a model that ultimately resembles a human language processor, achieve higher psychometric predictive power. Dynamic memory curricula, which aim to mimic the trajectory of human language development, instead can lead to stronger grammatical competence.
This interpretation is, in fact, consistent with the original ``Less is More'' hypothesis \citep{newport1990maturational}, which was proposed to explain how children acquire abstract grammatical generalizations.


%


\section{Discussion}


Overall, our results are partially in line with previous findings in the literature.
Consistent with the results in \citet{clark2025linear} and \citet{de-varda-marelli-2024-locally}, we find that a distance-based attention bias can lead to greater cognitive fit over an unbiased model, but only in some settings (BabyLM-10M 4-layer and Pile-2B 4-layer), and these differences are not statistically significant.
However, unlike \citet{mita2025developmentally}, we do not find gains on grammatical competence from a dynamic version of ALiBi; instead, such gains emerge only for the Less-to-More Forgetting Transformer, possibly reflecting differences in training data. 

Our first main contribution is to extend these previous studies across a wider range of attention mechanisms, model sizes, and training corpora.
When we do so, we find that the effects reported in prior work are not uniform across configurations, and rarely different from each other as per our statistical tests.
Rather, the gains from a distance-based bias are highly susceptible to model architecture and training data configurations, and in many cases, distance-biased models are worse for predicting reading times than vanilla Transformers.
These results suggest that conclusions from computational psycholinguistic modeling should be treated with caution without extensive validation.

Despite this overall null result, we do observe interesting trends in our data, which warrant further study.
For static models, content-based memory constraints tend to outperform distance-based constraints in predicting human reading times. 
The Forgetting Transformer outperforms ALiBi in every setting, and Stick-breaking attention outperforms ALiBi in four of six settings; however, these differences are not significant in our second-level analysis, so conclusions should be considered preliminary.
Our results suggest that intervener-based accounts may provide a more accurate characterization of the memory mechanisms underlying human language processing than distance-based decay accounts.
However, we do not interpret these results as providing evidence against \emph{cognitive} theories that have been traditionally modeled with distance-based memory decay, such as lossy-context surprisal \citep{futrell2020lossy}.
Rather, they suggest that such theories could and should be reworked to rely on intervener-based decay, rather than relying on distance-based decay alone.

Our third main finding is a dissociation between grammatical competence and psychometric predictive power.
A similar dissociation has been found, e.g., by \citet{steuer-etal-2023-large}, also looking at BabyLM-type models.
This suggests that the goals of modeling real-time sentence processing and modeling language acquisition may require fundamentally different approaches. Static constraints may be better suited for psychometric modeling, whereas a dynamic curriculum, which mimics a developmental trajectory, can instead lead to stronger grammatical competence, at least for the Forgetting Transformer.
These results raise questions about the possibility of finding a one-size-fits-all cognitive model, at least using autoregressive Transformers.
Rather, different modeling architectures may be needed to model different linguistic mechanisms.
A similar conclusion was recently reached by \citet{kuribayashi2026dual}, albeit for different types of real-time processing phenomena.

\section{Conclusion}

We set out to investigate which types of memory constraints in Transformer language models best capture human sentence processing behavior, and whether dynamic memory curricula improve cognitive fit over static constraints. 
We find that content-based constraints, particularly the Forgetting Transformer, tend to yield the strongest alignment with human reading time data, outperforming distance-based constraints across the majority of training configurations, although these differences are not statistically significant.
For dynamic memory curricula, the picture is more nuanced: dynamic models do not improve psychometric fit over static models, but the Less-to-More Forgetting Transformer shows stronger grammatical competence, revealing a dissociation between modeling real-time processing and modeling language acquisition.
Within the dynamic regime, the Less-to-More schedule yields better psychometric fit than the More-to-Less schedule, in line with the Less is More hypothesis.
These results underscore the value of systematically testing multiple cognitive theories across different modeling conditions, and suggest that intervener-based memory mechanisms deserve further attention in future computational modeling work.

\section*{Limitations}

This study has several limitations. First, the attention mechanisms we test are proxies for cognitive theories, not direct implementations. While we have argued for plausible mappings between each mechanism and a corresponding theory, other implementations are possible and may yield different results. Second, our evaluation is limited to English reading time corpora, and it remains an open question whether our findings generalize to other languages. Third, static models are evaluated only at the final training checkpoint (epoch 10), whereas dynamic models are evaluated at every epoch. This makes direct comparisons between static and dynamic BLiMP scores difficult, as grammatical competence may peak before the final epoch for all models.

\section*{Use of AI Assistants}
AI assistants were used during the preparation of this work for code development and editing support. 
Writing and code were subsequently checked, validated, and edited.
All scientific content, analysis, conclusions, and of course mistakes, are the authors' own.

\bibliography{custom}

\appendix

\section{Dynamic Model Implementation} \label{app:dynamic_models}

This appendix gives more information about the implementation of our two dynamic models.

\paragraph{Dynamic ALiBi}
We scale each fixed head-specific slopes of static ALiBi by a shared factor that changes across epochs:
\begin{align}
    a_{\i\j} &= \mathrm{softmax}_{\j}\left(\frac{\queries{\i}^\top \keys_{\j}}{\sqrt{\dim}} - \slopet \cdot \headslope(\i-\j)\right)
\end{align}
where $\headslope$ is the original head-specific slope, $\slopezero$ is the initial scaling factor, and $\decayrate$ is a scaling rate. In the Less-to-More condition, $\decayrate < 1$ (we use $\decayrate = 0.6$ following the original setting in \citealp{mita2025developmentally}), so $\slopet$ decreases over epochs, gradually relaxing the distance penalty. In the More-to-Less condition, $\decayrate > 1$ (we use $\slopezero = 0.0101$, $\decayrate = 1.6667$), so $\slopet$ increases over epochs, gradually imposing the penalty.

\paragraph{Dynamic Forgetting Transformer}
We scale the forgetting bias by a parameter \lamt that controls the strength of the content-based constraint and changes across training epochs:
\begin{align}
    a_{\i\j} &= \mathrm{softmax}_{\j}\left(\frac{\queries{\i}^\top \keys_{\j}}{\sqrt{\dim}} + \lamt \sum_{\k=\j+1}^{\i} \log f_\k\right)
\end{align}
where $\lamzero$ is the initial scaling factor and $\decayrate$ is a scaling rate. In the Less-to-More condition, $\decayrate < 1$ (we use $\decayrate = 0.6$ following \citeauthor{mita2025developmentally}), so $\lamt$ decreases over epochs, gradually relaxing the constraint. In the More-to-Less condition, $\decayrate > 1$ (we use $\lamzero = 0.0101$, $\decayrate = 1.6667$), so $\lamt$ increases over epochs, gradually imposing the constraint.


\section{Training Hyperparameters}
\label{sec:hyperparams}

All models are based on the OPT architecture \citep{zhang2022opt} and trained from scratch. We use two model sizes: a 2-layer model with 256 hidden dimensions and 4 attention heads, and a 4-layer model with 384 hidden dimensions and 6 attention heads. All models are trained for 10 epochs using the AdamW optimizer with a learning rate of 0.001, weight decay of 0.1, and betas of (0.9, 0.95). We use a cosine learning rate schedule with 1\% warmup. The context length is 512 tokens for BabyLM-10M and 1024 tokens for BabyLM-100M and Pile-2B. All models are trained with fp16 mixed precision and 3 random seeds (42, 43, 44). For dynamic models, the constraint parameter $\lambda$ is updated at the beginning of each epoch following the exponential decay schedule described in \Cref{sec:dynamic}. Full hyperparameters are summarized in \Cref{tab:hyperparams}.

\begin{table}[h]
\centering
\small
\begin{tabular}{lcc}
\toprule
\textbf{Hyperparameter} & \textbf{2-layer} & \textbf{4-layer} \\
\midrule
Hidden size & 256 & 384 \\
Attention heads & 4 & 6 \\
Learning rate & 0.001 & 0.001 \\
Optimizer & AdamW & AdamW \\
Betas & (0.9, 0.95) & (0.9, 0.95) \\
Weight decay & 0.1 & 0.1 \\
Warmup & 1\% of tokens & 1\% of tokens \\
LR schedule & cosine & cosine \\
Epochs & 10 & 10 \\
Precision & fp16 & fp16 \\
Seeds & 42, 43, 44 & 42, 43, 44 \\
\midrule
\multicolumn{3}{l}{\textit{Context length by corpus}} \\
\midrule
BabyLM-10M & 512 & 512 \\
BabyLM-100M & 1024 & 1024 \\
Pile-2B & 1024 & 1024 \\
\bottomrule
\end{tabular}
\caption{Training hyperparameters.}
\label{tab:hyperparams}
\end{table}

\section{Extended Training}
\label{app:ep20}

\begin{figure*}[t]
    \centering
    \includegraphics[width=0.8\textwidth]{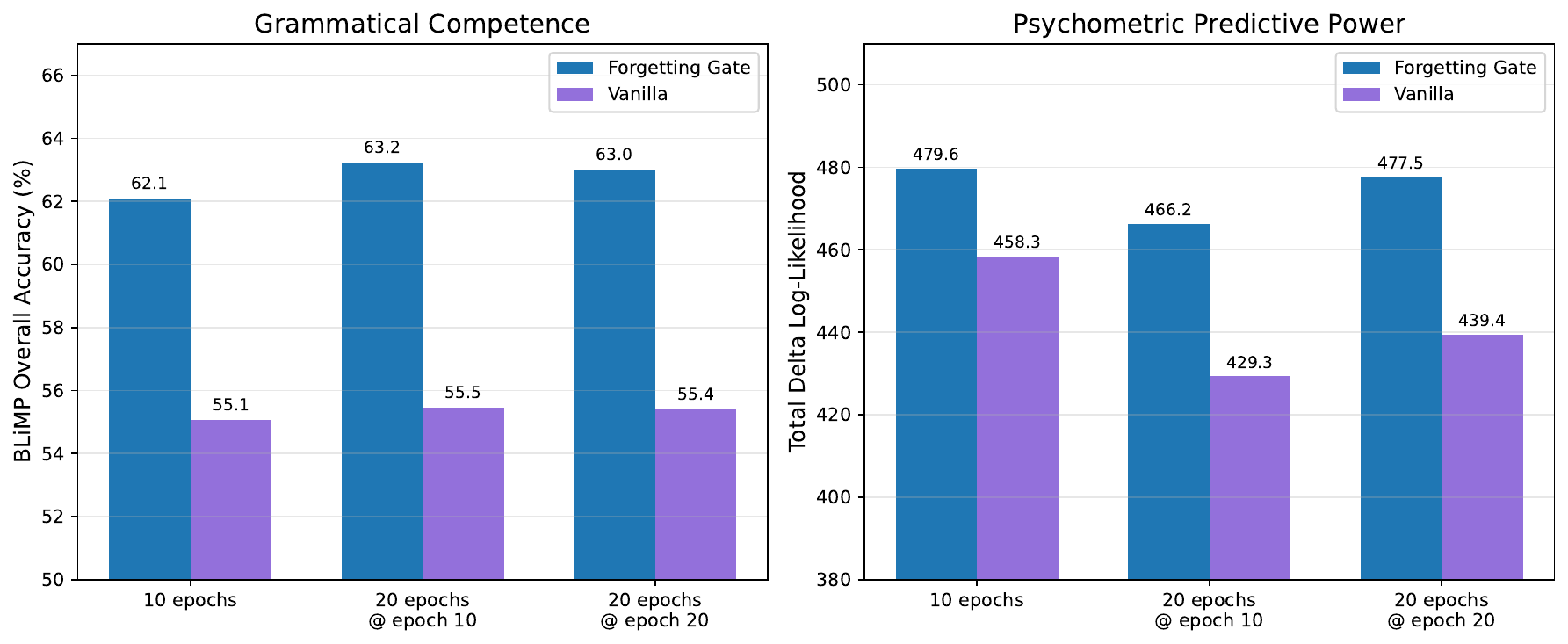}
    \caption{BLiMP accuracy (left) and \dll (right) for the Forgetting Transformer and vanilla model on BabyLM-10M (2-layer, seed 43), comparing the original 10-epoch models with a 20-epoch run evaluated at epochs 10 and 20.}
    \label{fig:ep20}
\end{figure*}

To test whether our models are undertrained at 10 epochs, we continued training the Forgetting Transformer and the vanilla model on BabyLM-10M (2-layer, seed 43) for 20 epochs. \Cref{fig:ep20} compares BLiMP accuracy and \dll for the original 10-epoch models with the 20-epoch models at epoch 10 and epoch 20. Note that the two runs differ at epoch 10 because the cosine learning rate schedule is stretched over the full training length.
We find little evidence of undertraining. BLiMP accuracy is essentially unchanged between epoch 10 and epoch 20 of the longer run (Forgetting Transformer: 63.2\% vs.\ 63.0\%; vanilla: 55.5\% vs.\ 55.4\%), and the 20-epoch models do not exceed the original 10-epoch models in \dll (Forgetting Transformer: 477.5 vs.\ 479.6; vanilla: 439.4 vs.\ 458.3). The Forgetting Transformer's advantage over the vanilla model also persists after extended training on both measures, suggesting that our main results are not an artifact of training length.

\section{Full Dynamic Results}
\label{app:dynamic_results_breakdown}

\begin{figure*}[t]
    \centering
    \includegraphics[width=\textwidth]{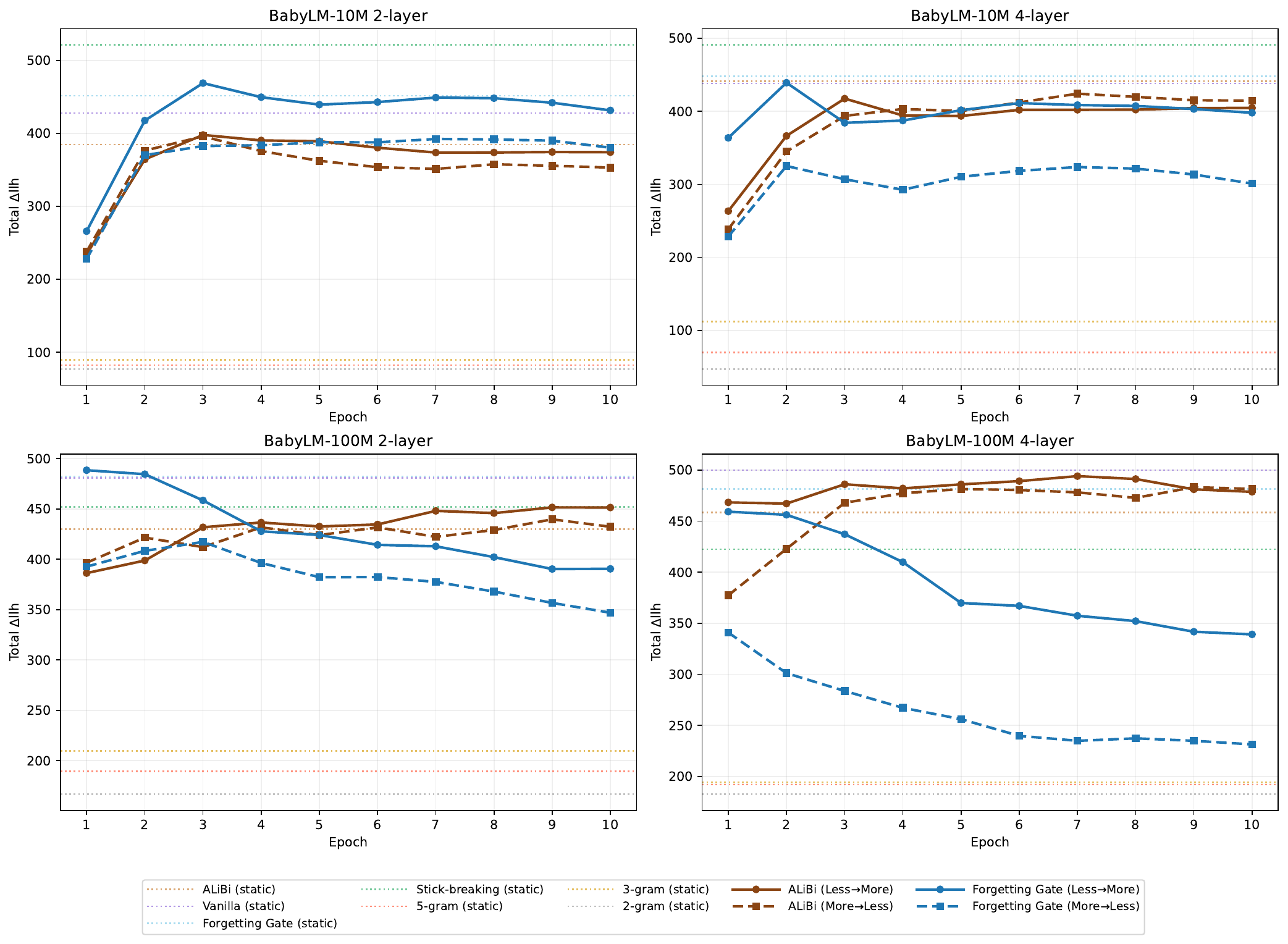}
    \caption{Per-setting \dll across training epochs for dynamic models.}
    \label{fig:appendix_dll}
\end{figure*}

\begin{figure*}[t]
    \centering
    \includegraphics[width=\textwidth]{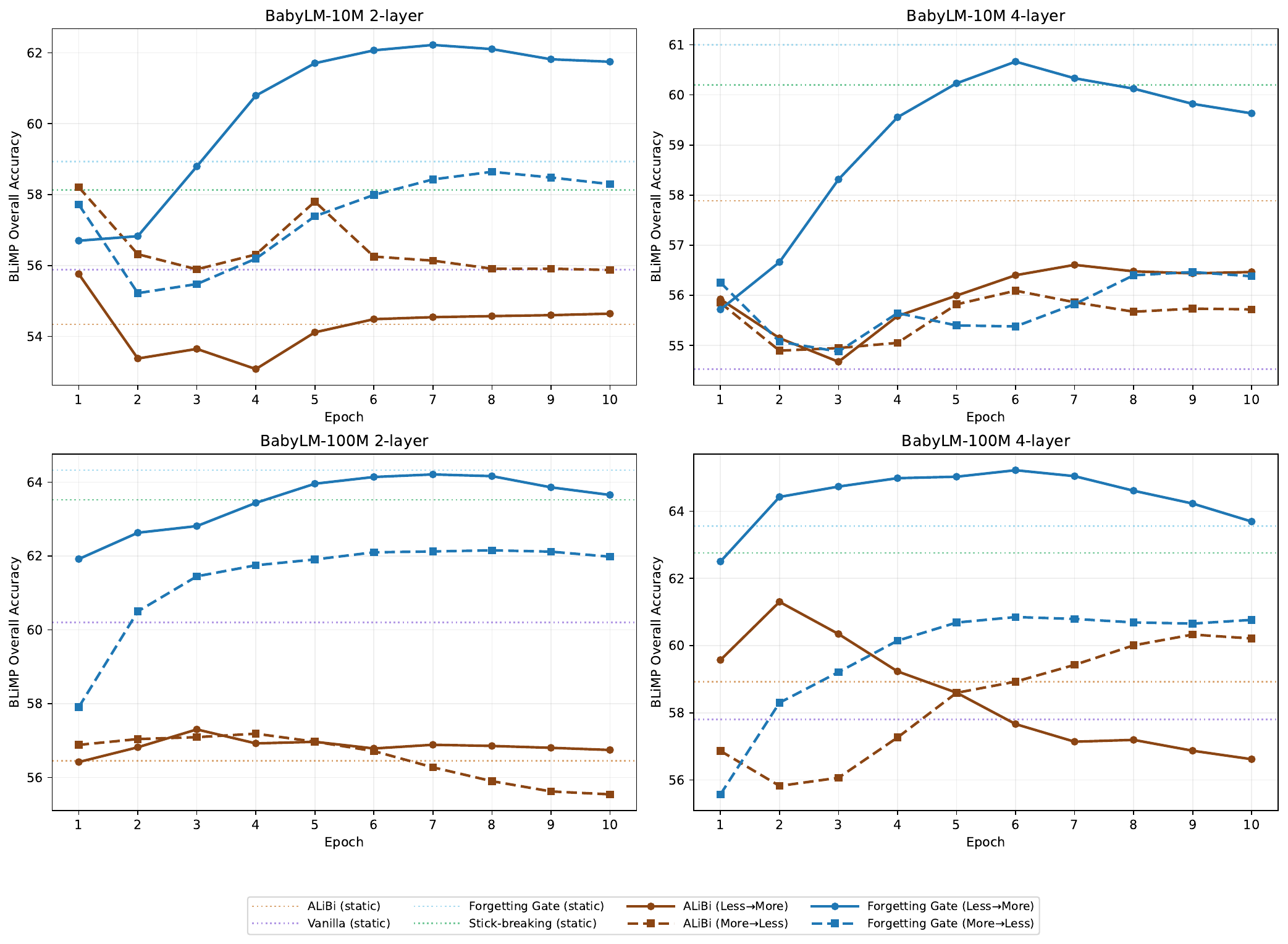}
    \caption{Per-setting BLiMP scores across training epochs for dynamic models.}
    \label{fig:appendix_blimp}
\end{figure*}

\Cref{fig:appendix_dll,fig:appendix_blimp} show per-setting \dll and BLiMP results for Experiment 2, broken down by training data size and model size.

\end{document}